\documentclass[letterpaper]{article} 
\usepackage{aaai25}  
\usepackage{times}  
\usepackage{helvet}  
\usepackage{courier}  
\usepackage[hyphens]{url}  
\usepackage{graphicx} 
\usepackage{natbib}  
\usepackage{caption} 
\usepackage{algorithm}
\usepackage{algorithmic}
\usepackage{pifont}   
\usepackage{amsthm,amsmath,amssymb}   
\usepackage[capitalize]{cleveref}   
\usepackage{multirow}   

\usepackage{newfloat}
\usepackage{listings}
\DeclareCaptionStyle{ruled}{labelfont=normalfont,labelsep=colon,strut=off} 
\floatstyle{ruled}
\newfloat{listing}{tb}{lst}{}
\floatname{listing}{Listing}
\title{Can Students Beyond The Teacher?\\ Distilling Knowledge from Teacher's Bias}
\author{
    Jianhua Zhang\textsuperscript{\rm 1},
    Yi Gao\textsuperscript{\rm 1}, 
    Ruyu Liu\textsuperscript{\rm 2, \rm 3}\thanks{Corresponding authors.}, 
    Xu Cheng\textsuperscript{\rm 1, \rm 3}\footnotemark[1], 
    Houxiang Zhang,\textsuperscript{\rm 4} 
    Shengyong Chen\textsuperscript{\rm 1}
}
\affiliations{
    \textsuperscript{\rm 1}School of Computer Science and Engineering, Tianjin University of Technology, Tianjin, 300384, China\\
    \textsuperscript{\rm 2}School of Information Science and Technology, Hangzhou Normal University, Hangzhou, 311121, China\\
    \textsuperscript{\rm 3}the Department of Technology, Management and Economics, Technical University of Denmark, Lyngby, Denmark \\
    \textsuperscript{\rm 4}Norwegian University of Science and Technology \\
    
    zjh@ieee.org,
    gaoyi01020304@stud.tjut.edu.cn,
    lry@hznu.edu.cn,
    xu.cheng@ieee.org,
    hozh@ntnu.no,
    sy@ieee.org
}

\usepackage{bibentry}

\begin{document}

\maketitle

\begin{abstract}
Knowledge distillation (KD) is a model compression technique that transfers knowledge from a large teacher model to a smaller student model to enhance its performance. Existing methods often assume that the student model is inherently inferior to the teacher model. However, we identify that the fundamental issue affecting student performance is the bias transferred by the teacher. Current KD frameworks transmit both right and wrong knowledge, introducing bias that misleads the student model. To address this issue, we propose a novel strategy to rectify bias and greatly improve the student model's performance. Our strategy involves three steps: First, we differentiate knowledge and design a bias elimination method to filter out biases, retaining only the right knowledge for the student model to learn. Next, we propose a bias rectification method to rectify the teacher model's wrong predictions, fundamentally addressing bias interference. The student model learns from both the right knowledge and the rectified biases, greatly improving its prediction accuracy. Additionally, we introduce a dynamic learning approach with a loss function that updates weights dynamically, allowing the student model to quickly learn right knowledge-based easy tasks initially and tackle hard tasks corresponding to biases later, greatly enhancing the student model's learning efficiency. To the best of our knowledge, this is the first strategy enabling the student model to surpass the teacher model. Experiments demonstrate that our strategy, as a plug-and-play module, is versatile across various mainstream KD frameworks. We will release our code after the paper is accepted.
\end{abstract}

%

\section{Introduction}
\label{sec:intro}

\begin{figure}[h]
\centering
\includegraphics[scale=0.45]{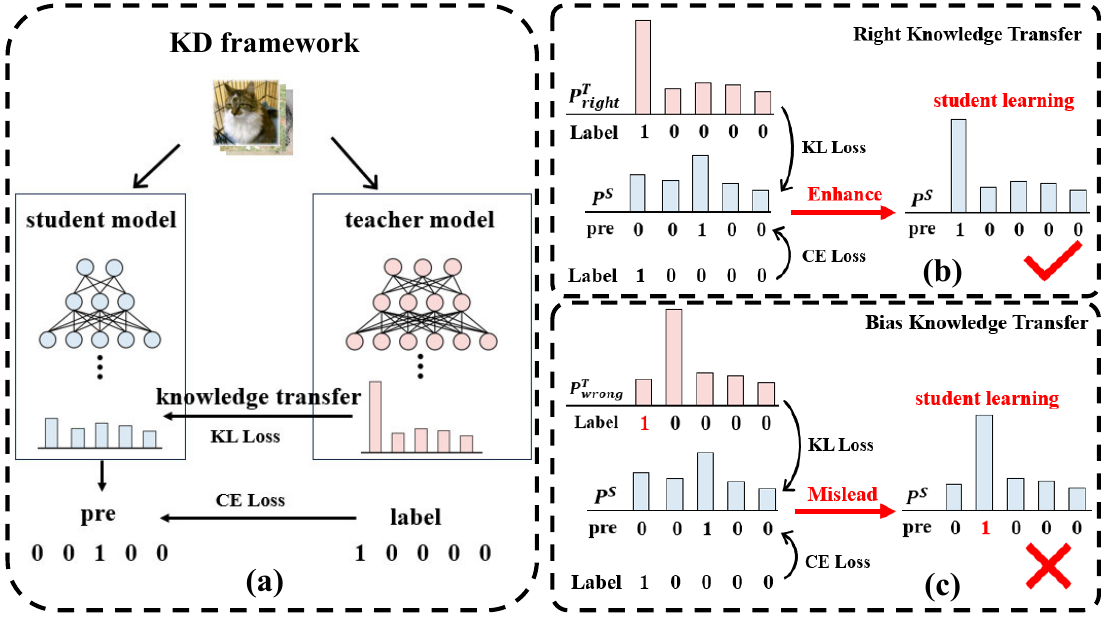}
\caption{Diagram of the knowledge transfer process in KD. (a) The KD framework based on a teacher-student network model. (b) Right knowledge transfer helps the student model learn effectively. (c) Bias knowledge transfer can mislead the student model with biased knowledge.}
\label{fig1-2}
\end{figure}

Knowledge Distillation (KD) was proposed by \cite{hinton2015distilling}. This method uses a pre-trained model as a teacher, from which useful knowledge is extracted and transferred to train a smaller student model. This process enables the student model to achieve performance close to that of the teacher model.
Consequently, KD effectively compresses models while maintaining their performance, making it widely applicable in scenarios that require both efficiency and accuracy, such as on mobile and embedded devices, real-time image processing, and video analysis. 
Despite efforts to improve student model performance through various KD methods \cite{park2019relational,tian2019contrastive,chen2021distilling,sarridis2022indistill,patel2023learning}, including those based on logit layers \cite{muller2019does,zhao2022decoupled} and intermediate feature layers \cite{park2022prune, sarridis2022indistill,LIANG2024110216,TANG2023109320,10.1145/3664647.3680591}, these approaches have consistently failed to surpass the teacher model. 
The fundamental reason is that these methods overlook the fact that the knowledge transmitted by the teacher is not entirely accurate. Bias from the teacher can misguide the student, thereby limiting the student model's performance. Moreover, as data types become more diverse and tasks become more complex, the impact of bias knowledge will further intensify.

We acknowledge the bias present in the KD framework. As shown in \cref{fig1-2}, the student model learns from the true labels and the teacher model's predictions through cross-entropy (CE) loss and Kullback-Leibler divergence (KL) loss, respectively. The student's learning is enhanced when the teacher's predictions ($P^T_{right}$) align with the true labels. Conversely, when the teacher's predictions ($P^T_{wrong}$) do not match the true labels, they mislead the student, causing its learning to deviate and resulting in wrong predictions. Based on these two observations, we categorize the knowledge transferred by the teacher into right knowledge and wrong knowledge, with the latter further defined as bias.

To mitigate the impact of teacher bias on student learning performance, we propose a novel bias rectification strategy. First, based on the categorization of right knowledge and bias, we design a bias elimination method that separates bias from the transferred knowledge, retaining only the right knowledge. This directly eliminates the impact of bias on the student model, thereby greatly enhancing its performance under the guidance of the right knowledge. Second, the student model cannot learn knowledge from the data resulting in teachers’ bias, if we directly remove those data when training student model, we further design a bias rectification method that utilizes weighted adjustment to convert bias into the right knowledge. This fundamentally addresses the impact of bias, allowing the student model to learn from both the right knowledge and the rectified bias, thereby greatly improving prediction accuracy. 
Third, we observe that the learning times for the student model for easy tasks based on right knowledge and for hard tasks corresponding to biases is much different. Therefore, we propose a dynamic learning approach with an improved loss function to optimize the model. Initially, the student focuses on learning easy tasks, and as learning capability increases, more harder tasks are gradually introduced in the later stages of training. This approach significantly reduces the KD time and greatly enhances the student model's prediction efficiency. Finally, we validate the effectiveness of our bias rectification strategy for KD frameworks on complex tasks like image classification and object detection. Extensive experiments demonstrate that our method outperforms state-of-the-art (SOTA) KD methods on benchmark datasets and can be used as a plug-and-play module to enhance existing KD frameworks. Moreover, it is the first time that the student model can surpass the teacher model in KD framework through the proposed strategy.

The main contributions of our paper are as follows:
\begin{itemize}
    \item Through in-depth theoretical analysis, we have demonstrated the presence of bias in KD and its detrimental impact on student model performance.
    \item We propose a novel bias rectification strategy through which the student model surpasses the teacher model for the first time. In this strategy, we not only eliminate errors to strengthen the transmission of correct knowledge but also rigorously rectify biases to mitigate the misleading effects of incorrect knowledge. 
    \item We propose a dynamic learning approach that allows the student model to quickly master easy tasks based on right knowledge in the early stages of training, while addressing hard tasks related to biases in the later stages. This approach significantly improves learning efficiency.
    \item We validated the effectiveness of our strategy by showcasing the student model's superior performance on two types of tasks across three benchmark datasets. Additionally, as a plug-and-play model, the strategy is versatile and can enhance existing KD frameworks.
\end{itemize}

\section{Related Work}
KD methods \cite{gou2021knowledge,wang2021knowledge,chen2022semantic} can be categorized into two categories: logits-based and features-based. The logits-based method primarily involves having the student learn from the teacher's soft labels. Initially proposed by Hinton et al.\cite{hinton2015distilling}, this method is known for its simplicity and ease of implementation, achieving significant results in the early stages of neural network training. Subsequent work by \cite{DBLP:conf/iclr/KimK17} introduced the use of class-distance loss to enhance knowledge transfer to the student model.
Zhao et al. \cite{zhao2022decoupled} improved the effectiveness of logits-based KD by decoupling probabilities in the logits. More recently, CTKD method \cite{li2023curriculum} has further enhanced student model performance by dynamically adjusting the temperature hyperparameter in KD. However, existing methods generally assume that all transferred knowledge is right, overlooking the bias introduced when the probability distribution predicted by the teacher model does not match the true labels. These biases are directly transferred to the student model through the logit layer via KL loss, which misleads the student learning the biased knowledge. This results in a performance ceiling that the student model cannot surpass.

More researchers \cite{paulin2012knowledge,huang2017like,ZHANG2021107659} have realized that the intermediate layer features of the teacher network contain valuable information. Therefore, features-based KD \cite{passalis2018learning,zhu2021complementary,park2021learning} has been widely explored to improve student performance by transferring knowledge from the teacher model's intermediate features. FitNet \cite{romero2014fitnets} is one of the first KD methods based on intermediate layers. It uses a small regression to align the intermediate layer features of the teacher and student model. Subsequent works \cite{park2019relational,tian2019contrastive,chen2021distilling,sarridis2022indistill} have delved into in-depth research on effectively transmitting the teacher's intermediate layer features to the student. 
Current methods \cite{XU2023109338,CHO2023109541,ZHU2024110545,XIE2024110455} primarily focus on effectively extracting and transferring knowledge from the intermediate feature layers of the teacher model without considering the rightness of the transferred knowledge. Intermediate features often contain rich semantic information, but biased knowledge is hidden among numerous neurons, making it difficult to detect. This biased knowledge subtly affects the accuracy of semantic information extraction, leading to deviations in final predictions. Therefore, transferring such biased knowledge to the student model can also mislead its final predictions. 

While the previous method \cite{zhou2020channel} have attempted to ensure the rightness of feature transfer by simply removing the intermediate features of wrongly predicted samples, this approach of directly associating wrong predictions with intermediate features is unreasonable. Intermediate features of right samples may also contain biased knowledge, Conversely, wrong samples may contain a lot of right feature information, and removing all of them affects the completeness of feature extraction. Although adversarial defense methods \cite{10679891} can effectively mitigate the impact of biased intermediate features, these methods still cannot effectively eliminate the bias in the teacher model.

\begin{figure*}[ht]
\centering
\includegraphics[scale=0.74]{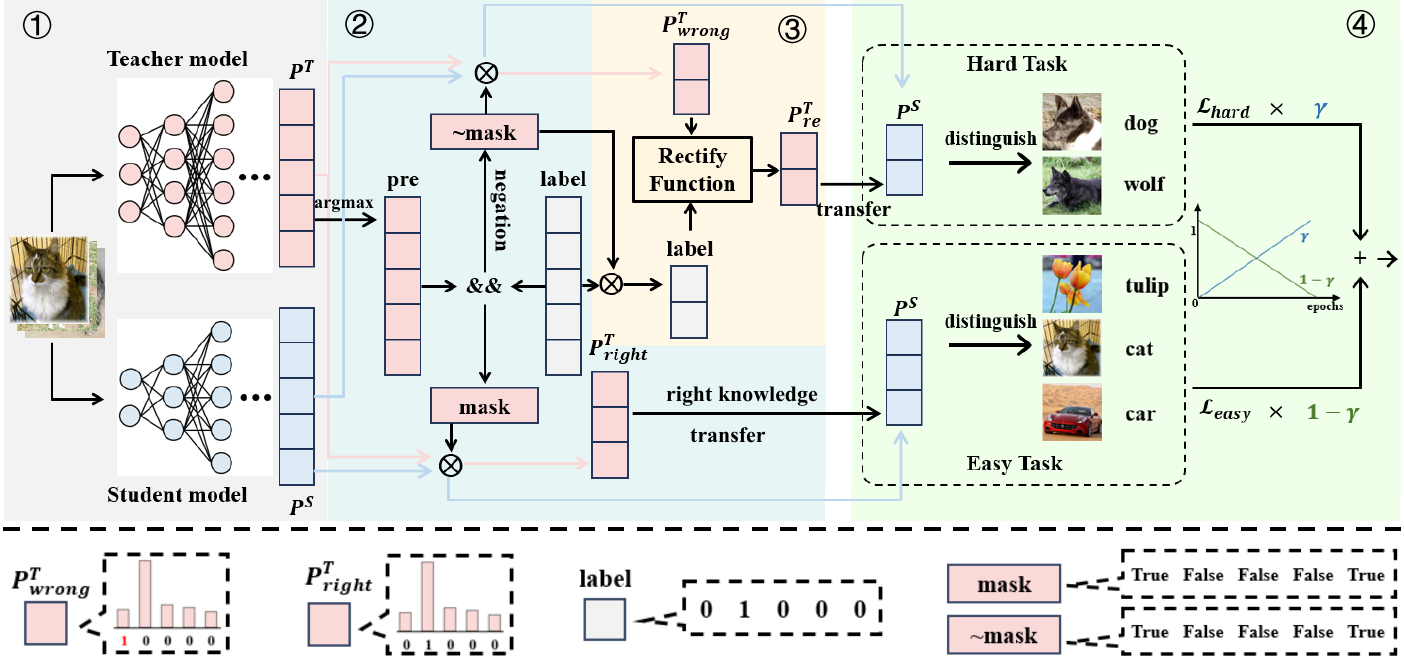}
\caption{The overall framework of our framework. Module \ding{172} is a KD framework based on the teacher-student model. Module \ding{173} is the bias elimination method. Module \ding{174} is the bias rectification method. Module \ding{175}  is the dynamic learning approach.
}
\label{fig1}
\end{figure*}

\section{ Methodology }
\label{Section3}

\subsection{Definitions}
We first define and introduce several key concepts and terms that will be used throughout this section.
\textbf{Right and biased knowledge}: right knowledge comes from cases where the teacher's predicted probability distribution aligns with the true labels. Biased knowledge, or wrong knowledge, arises when the teacher's predicted probability distribution does not match the true labels.
\textbf{Easy and Hard Tasks}: For the student, tasks where the teacher can make right predictions are considered easy tasks, and the knowledge transferred by the teacher is right knowledge. Conversely, tasks where the teacher makes wrong predictions are considered hard tasks, and the knowledge transferred is biased knowledge.

\subsection{Rethinking Knowledge Transmission}
\label{Section3.1}

In the KD framework, the loss is typically composed of KL loss and CE loss, with the former used to constrain the student's learning of the knowledge transferred by the teacher and the latter to constrain the student's learning of the label. We use $T$ and $S$ to represent the predicted probability values of the teacher and the student, respectively, and $Y$ to represent the label. $t_i \in T$, $s_i \in S$ and $y_i \in Y$ $i \in n$. Generally, we use
\begin{align}
    Loss &= \mathcal{L}_{KL}\left( T,S \right) + \mathcal{L}_{CE}\left( Y,S \right)  \nonumber  \\
    &= \sum\limits_{i=1}^n{t_i{ln\frac{t_i}{s_i}}} + \left(-\sum_{i=1}^n {y_i ln s_i} \right)
\end{align}
as the loss function. 
When minimizing the loss function through gradient backpropagation, minimizing $KL$ will make the $s_i$ approach $t_i$, and minimizing $CE$ will make the $s_i$ approach $y_i$. Let $a$ and $b$ represent two categories, where $t_a$ and $t_b$ are the predicted probabilities of the teacher, $s_a$ and $s_b$ are the predicted probabilities of the student, and $y_a$ and $y_b$ are the corresponding class labels. Assuming $y_a = 1$ and $y_b = 0$, when minimizing the loss function, we have
\begin{equation}
    min\left(Loss\right) \Rightarrow
    \begin{cases}
     s_a \to 1~,~~~~ s_a \to t_a~;\\
     s_b \to 0~,~~~~ s_b \to t_b~.
    \end{cases}
\end{equation}

If the teacher's predictions match the labels (i.e., $t_a\to1,t_b\to0$), then it is beneficial for the student to learn from teacher, where $ s_a \to t_a,s_b\to t_b$, and consequently $ s_a \to 1$ and $s_b \to 0$.
However, when the teacher's predictions do not match the labels (i.e., $  t_a\to0,t_b\to1$), teaching the student to learn from teacher, where $ s_a \to t_a,s_b\to t_b$, actually results in $s_a \to 1 $ and $s_b \to 0 $, which is misleading for the student.

In fact, since the teacher model's own prediction accuracy is not guaranteed to be perfect, the bias inevitably misleads the student. Therefore, we demonstrate that knowledge can be divided into right knowledge and bias, with the bias having a negative impact on the performance of the student model.

\subsection{Eliminating Biased Knowledge from Teacher}
\label{section3.2}

To further eliminate biased knowledge in KD and ensure the rightness of knowledge transferred from the teacher modelk, we design a eliminate module, as depicted in the blue region in \cref{fig1}. Firstly, we convert the predicted values output by the teacher model into corresponding predicted probabilities, recorded as $P^T$.
We use the $argmax$ operation to convert the teacher's predicted probabilities $P^T$ into a 0-1 vector $pre$. This vector is then compared to the true labels using a logical $AND$ operation, marking them as True if they match and False if they do not. We store these True and False values in the mask table in the order corresponding to the teacher's prediction index to record the rightness of the teacher's information. 
We classify the $P^T$ into right knowledge (recorded as True in the mask table) and biased knowledge (recorded as False in the mask table). Thus far, we have distinguished right and biased knowledge in the teacher model and stored their indexes in the mask table. This enables students to later learn the knowledge transferred from the teacher under these indexes and perform gradient updates only using the corresponding knowledge in each batch of training.

We obtain the right knowledge by multiplying $P^T$ and $P^S$ with the mask. By inverting the mask to get $\sim$mask, we can then multiply $P^T$ and $P^S$ with $\sim$mask to obtain the biased knowledge.

\begin{figure*}[ht]
\centering
\includegraphics[scale=0.8]{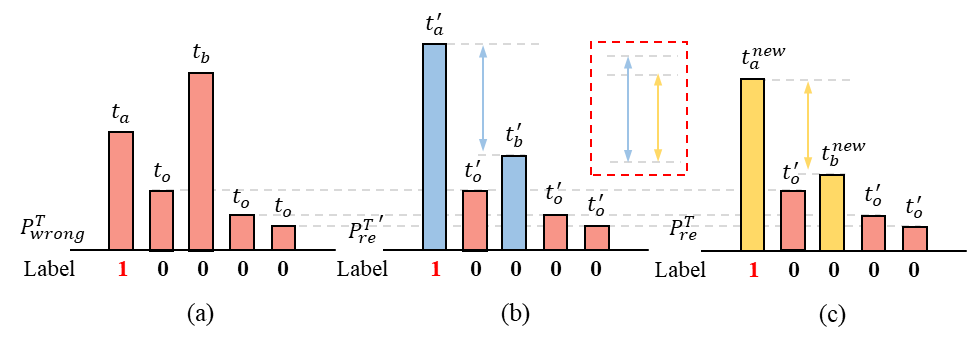}
\caption{Diagram of the rectification method. Each column corresponds to a category, $Label$ corresponds to the one-hot vector of ground truth, $P^T_{wrong}$ corresponds to the probability value predicted by the teacher on each category, $P^T_{re}$ corresponds to the probability value after rectification. (a) shows the teacher's wrong prediction, (b) shows the predicted probability is consistent with $Label$ after rectification, (c) represents further adjustment of the rectification predicted probabilities.
}
\label{fig2}
\end{figure*}

\subsection{Rectifying Biased Knowledge from Teacher}
\label{section3.3}

Eliminating bias can improve the overall predictive performance of the student model to some extent. However, since eliminating bias essentially removes the wrong predictions, it does not enhance the student model's ability to learn these hard tasks. As the data becomes more diverse and complex, the amount of bias will correspondingly increase, leading to a performance bottleneck in enhancing the student model through bias elimination. Therefore, we further propose a bias rectification method (as shown in Module \ding{174} of \cref{fig1}) aimed at rectifying the teacher model's wrong predictions, thereby equipping the student model with the ability to learn hard tasks and further improving its predictive performance.
Suppose $t_a$ represents the predicted probability of the teacher corresponding to the label $y_a$ as 1, $t_b$ represents the largest predicted probability by the teacher, $y_b$ represents the label value corresponding to $t_b$, $\mathbf{t_o}=\{t_{oi}\}$ represent the others probability prediction values, $\mathbf{y_o}=\{y_{oi}\}$ represents the others label values. $y_a$, $y_b$ and $\mathbf{y_o}$ sum to 1, as shown in \cref{fig2} (a). When we take the bias prior knowledge of the teacher, there is the following relationship:
\begin{equation}
\label{eq3}
t_a + t_b + \mathbf{t_o} = 1 ,~~
 \begin{cases}
 y_a = 1 ,t_a \to 0~;  \\
 y_b = 0 ,t_b \to 1~;  \\
 \mathbf{y_o} = \{\mathbf{0}\} ,\mathbf{t_o} \to \{\mathbf{0}\}~.
\end{cases} 
\end{equation}

In \eqref{eq3}, $t_a$ and $t_b$ are opposite to the corresponding labels $y_a$ and $y_b$, as we discussed, such biased knowledge from the teacher is harmful. We have considered the following requirements for rectifying this bias. First, we aim to adjust the values of $t_a$ and $t_b$ to be consistent with $y_a$ and $y_b$ . Second, we ensure that the sum of all adjusted $t_i$ values is 1 (satisfying the principle of probability distribution). Finally, since the $\mathbf{t_o}$ represents other classes unrelated to the two classes that need rectification, we cannot affect $\mathbf{t_o}$ when adjusting $t_a$ and $t_b$.
Based on these considerations, we propose the following strategy.

First, we weight $t_a$ and $t_b$ with the corresponding labels $y_a$ and $y_b$. Then, we take the average to obtain the new probabilities $t^{'}$, as shown in the following equations:

\begin{equation}
\begin{cases}
    t_a^{'} = \frac{t_a + y_a}{2} = \frac{t_a+1}{2} > 0.5,   \\
    t_b^{'} = \frac{t_b + y_b}{2} = \frac{t_b}{2}  < 0.5,  \\
    \mathbf{t_o^{'}} = \mathbf{t_o}
\end{cases}
\end{equation}

The rectified probabilities obtained are shown in \cref{fig2}(b), where the largest probability value corresponds to a label value of 1, indicating that the predicted value correctly corresponds to the label value. This transformation from biased knowledge to right knowledge is illustrated.

Because
\begin{align}
    &t_a^{'} + t_b^{'}=\frac{t_a+t_b+y_a+y_b}{2} = \frac{t_a+t_b+1}{2} , \nonumber   \\
    &t_a + t_b < 1 ,
\end{align}
so 
\begin{equation}
    \frac{t_a+t_b+1}{2} > t_a+t_b \Rightarrow t_a^{'} + t_b^{'} + \mathbf{t_o^{'}} > 1
\end{equation}
Considering that the sum of the probabilities for the new $t_a^{'}$, $t_b^{'}$ and $\mathbf{t_o^{'}}$ is not equal to 1, disrupting the principle of probability distribution.
Therefore, we need to readjust the value of $t_a^{'}$ and $t_b^{'}$ to make $t_a^{'} + t_b^{'} + \mathbf{t_o^{'}} = 1$ without changing $\mathbf{t_o^{'}}$. Let $t^{new}$ represent the adjusted value, then there is the following formula:

\begin{equation}
 t_a^{new} = t_a^{'} {\times} \frac{t_a + t_b}{t_a^{'} + t_b^{'}} ,~~~~
 t_b^{new} = t_b^{'} {\times} \frac{t_a + t_b}{t_a^{'} + t_b^{'}}
\end{equation}   
At this point, we have obtained the final rectification probability values (shown in  \cref{fig2}(c)), ensuring that the probabilities sum to one without altering the value of $\mathbf{t_o}$, while also ensuring that the new predicted results align with the label.

\subsection{Dynamic Learning Approach}
Since the student's learning efficiency differs between easy and hard tasks, we consider focusing on easy tasks in the early stages of training. As the student's capability improves, we increase the emphasis on tackling hard tasks in the later stages of training (as shown in Module \ding{175} of \cref{fig1}). Both right knowledge and rectified knowledge play important roles in the student learning process and contribute to the student's eventual surpassing of the teacher. Consequently, we have defined two constraint functions to respectively constrain the student's learning of these two types of knowledge.
\begin{align}
    \mathcal L_{easy} &= \mathcal L_{KL}\left(P^{S_{r}} \| P^{T_{r}}\right), \nonumber    \\ 
    \mathcal L_{hard} &= \mathcal L_{KL}\left(P^{S_{re}} \| P^{T_{re}}\right)
\end{align}

$\mathcal{L}_{easy}$ represents the loss function for transmitting easy task knowledge, where $P^{T_{r}}$ denotes the probability distribution of the right knowledge in the teacher and $P^{S_{r}}$ represents the probability distribution of the corresponding student. We use the $KL$ divergence to constrain the approximation between the two distributions. $\mathcal{L}_{hard}$ represents the loss function for transmitting hard task knowledge, where $P^{T_{re}}$ represents the probability distribution of knowledge after rectification in the teacher and $P^{S_{re}}$ represents the probability distribution of the corresponding student. Similarly, we use the $KL$ divergence to constrain the approximation between the two distributions.

The overall loss function is defined as follows:
\begin{equation}
\label{equation2}
    \mathcal{L}_{all} = \left( 1-\gamma \right) \left( \mathcal L_{CE} +\mathcal{L}_{easy} \right) + \gamma  \mathcal{L}_{hard}
\end{equation}
In order to shorten the training cycles, we employed a method for dynamically adjusting the student's learning focus. We have the following dynamic adjustment coefficient as $\gamma = \frac{e}{E}$. $e$ represents the current training iteration, and $E$ denotes the total training iterations.

By adjusting $\gamma$ throughout the entire training process, the student is initially encouraged to prioritize learning easy tasks as the training iterations progress. This approach allows the student to quickly attain a foundational level of proficiency in basic knowledge. Subsequently, with the deepening of training, the focus gradually shifts towards the learning of more challenging knowledge. Effective fine-tuning on the established foundational proficiency leads to better performance enhancements. Ultimately, this strategy aims to achieve the goal of reducing the training time cost and improving the convergence effectiveness.

\begin{table*}
\centering
\scalebox{0.7}{
\begin{tabular}{c|ccccc|cccc}
\hline
\multirow{15}{*}{\rotatebox{90}{CIFAR-10 dataset}}  
& \multirow{2}{*}{teacher} & ResNet-50 & ResNet-101 & WRN-40-2 & VGG-13 & ResNet-50   & VGG-13   & ResNet-50     & WRN-40-2  \\
&       & 96.08     & 96.83   & 93.52  &93.28   & 96.08  & 93.28  & 96.08  & 93.52      \\
& \multirow{2}{*}{student} & ResNet-18 & ResNet-34  & WRN-16-2 & VGG-8  & MobileNet-V2 & MobileNet-V2 & ShuffleNet-V1 & ShuffleNet-V1\\
&      & 93.30     & 94.59  & 90.50    & 88.78    &90.08    &90.08         &90.70   &90.70             \\ \cline{2-10} 
& KD\cite{hinton2015distilling}     & 93.81   &94.87      & 92.73    &89.30   &93.79      &91.73         &91.21    &91.09          \\
& FitNet\cite{romero2014fitnets} & 93.87   &94.80      & 92.01    &89.70   &93.70      &91.79         &91.30          &90.97          \\
& RKD\cite{park2019relational}    & 94.37   &95.01      & 92.79    &89.93   &94.01      &92.07         &92.47          &91.21          \\
& CRD\cite{tian2019contrastive}    & 94.90   &95.39      & 93.01    &90.74   &94.21      &92.37         &92.80          &91.30          \\
& OFD\cite{heo2019comprehensive}   & 94.87    &95.20      & 92.97    &90.51   &94.22      &92.20         &92.59          &91.24          \\
& ReviewKD\cite{chen2021distilling}   & 95.07   &95.57   & 93.17   &91.99   &95.08      &92.90         &93.97          &92.68          \\
& Indistill\cite{sarridis2022indistill}   & 94.99   &95.19   & 93.20  &91.63   &94.38      &92.43         &93.50          &92.06          \\
& DKD\cite{zhao2022decoupled}    & 95.02   &95.49      & 93.21    &91.75   &94.79      &92.81         &93.89          &92.60          \\
& CTKD\cite{li2023curriculum}   & 94.73    &95.24     & 93.01    &91.10   &94.20      &92.05         &93.41          &92.19          \\
&\textbf{Ours}    &\textbf{96.89} &\textbf{96.99} & \textbf{94.18}   &\textbf{93.60}   &\textbf{96.28}  &\textbf{93.76} &\textbf{96.20} &\textbf{93.70}           \\
& $\Delta$    & +1.82   &+1.42     & +0.97    & +1.61    &+1.20   &+0.86    & +2.23    &+1.02               \\ \hline
\multirow{15}{*}{\rotatebox{90}{CIFAR-100 dataset}}  
& \multirow{2}{*}{teacher} & ResNet-50 & ResNet-101 & WRN-40-2 & VGG-13 & ResNet-50    & VGG-13       & ResNet-50     & WRN-40-2      \\
 &   & 79.34  & 81.87  & 73.57  & 74.64  & 79.34  & 74.64   & 79.34  & 73.57    \\
& \multirow{2}{*}{student} & ResNet-18 & ResNet-34  & WRN-16-2 & VGG-8   & MobileNet-V2 & MobileNet-V2 & ShuffleNet-V1 & ShuffleNet-V1 \\
 &   & 69.75  &72.50   & 70.26  & 70.36  & 65.40  & 65.40   & 70.50  & 70.50    \\  \cline{2-10}
& KD\cite{hinton2015distilling}        & 71.56 & 73.17   & 70.92    & 72.98  & 69.35  & 67.37   &71.97               & 70.83           \\
& FitNet\cite{romero2014fitnets}    & 70.21 & 73.08   & 70.98    & 71.02  & 65.56  & 64.14   &71.03               & 70.73              \\
& RKD\cite{park2019relational}       & 71.67 & 73.87   & 71.32    & 71.48  & 66.73  & 64.52   &72.84               & 71.21         \\
& CRD\cite{tian2019contrastive}       & 72.16 & 74.60   & 71.37    & 73.94  & 71.11  & 69.73   &73.10               & 72.05          \\
& OFD\cite{heo2019comprehensive}       & 71.98 & 73.91   & 71.10    & 73.95  & 71.04  & 69.48   &72.78               & 71.85        \\
& ReviewKD\cite{chen2021distilling}  & 73.19 & 75.80   & 71.59    & 74.84  & 72.89  & 70.37   &76.10          & 73.14     \\
& Indistill\cite{sarridis2022indistill}  & 73.17 & 75.17   & 71.09    & 74.65  & 72.36  & 70.01   &75.48          & 72.10         \\
& DKD\cite{zhao2022decoupled}       & 73.97 & 75.67   & 71.54    & 74.68  & 72.35  & 69.71   &75.88               & 73.10            \\
& CTKD\cite{li2023curriculum}      & 72.29 & 74.58   & 71.45    & 73.52  & 70.46  & 68.46   &75.34               & 71.78              \\
&\textbf{Ours}   &\textbf{81.50}  &\textbf{84.77}  & \textbf{74.30} &\textbf{78.03}  &\textbf{80.45}  &\textbf{75.71}  &\textbf{81.07}                       & \textbf{74.78}              \\
& $\Delta$   & +7.53   & +8.97    & +2.71     &+3.19    & +8.10   & +5.34   &+4.97            &+1.64   \\  \hline           
\end{tabular}}
\caption{The results on the CIFAR-10 and CIFAR-100 dataset. $\Delta$ represents the classification accuracy improvement over the best result of the current SOTA methods in knowledge distillation. }
\label{Table1}
\end{table*}

\section{Experiment}
We conducted experiments on three classification datasets, \textbf{CIFAR-10}\cite{krizhevsky2009learning}, \textbf{CIFAR-100}\cite{krizhevsky2009learning}, and \textbf{ImageNet 1K}\cite{russakovsky2015imagenet}, as well as on an object detection dataset \textbf{MS-COCO}\cite{lin2014microsoft}.

\subsection{Comparison with SOTA KD methods}
\textbf{Classification on CIFAR dataset.} In \cref{Table1}, we present the results of KD in two datasets.
Our method demonstrates superior performance over the teacher on both the CIFAR-10 and CIFAR-100 datasets. Due to the increased difficulty of the 100-class classification task compared to the 10-class task, the teacher model makes more wrong predictions, resulting in more bias being transmitted. As a result, our method demonstrates superior performance.

For cross-model KD, we employed MobileNet \cite{sandler2018mobilenetv2} and ShuffleNet \cite{zhang2018shufflenet} as student models, both exhibiting significant structural and parameter differences compared to the teacher models (ResNet-50 \cite{he2016deep}, VGG \cite{simonyan2014very} and WRN \cite{zagoruyko2016wide}). This is sufficient to validate the capability of cross-model knowledge distillation. In this scenario, our method still achieves the best performance, enabling the student model to surpass the teacher model. Particularly in the challenging CIFAR-100 task, our method shows the most significant improvement compared to the SOTA methods.

\begin{table}[]
\centering
\scalebox{0.7}{
\begin{tabular}{cc|c|c}
\hline
    &           & top-1 acc(\%) & top-5 acc(\%) \\ \hline
\multicolumn{1}{c|}{}     &Teacher: ResNet-34 & 73.31         & 91.42         \\
\multicolumn{1}{c|}{}     &Student: ResNet-18 & 69.75         & 89.07         \\ \hline
\multicolumn{1}{c|}{\multirow{5}{*}{features}} & AT\cite{zagoruyko2016paying}        & 70.69         & 90.01         \\
\multicolumn{1}{c|}{}    & OFD\cite{heo2019comprehensive}    & 70.81         & 89.98         \\
\multicolumn{1}{c|}{}    & CRD\cite{tian2019contrastive}     & 71.17         & 90.13         \\
\multicolumn{1}{c|}{}   & ReviewKD\cite{chen2021distilling}  & 71.61         & 90.51         \\
\multicolumn{1}{c|}{}   & InDistill\cite{sarridis2022indistill} & 71.63         & 90.37         \\ \hline
\multicolumn{1}{c|}{\multirow{4}{*}{logits}}   & KD\cite{hinton2015distilling}      & 71.03         & 90.05         \\
\multicolumn{1}{c|}{}    & DKD\cite{zhao2022decoupled}    & 71.70         & 90.41         \\
\multicolumn{1}{c|}{}     & CTKD\cite{li2023curriculum}      & 71.32         & 90.27         \\
\multicolumn{1}{c|}{}                          & \textbf{Ours}      & \textbf{73.38}         & \textbf{91.71}         \\ 
\multicolumn{1}{c|}{}                          & {$\Delta$}      & +1.75    & +1.2      \\  \hline
\end{tabular}}
\caption{acc refers to classification accuracy($\%$) on the ImageNet 1K dataset, top-1 and top-5 are standards for calculating classification accuracy on the validation set. $\Delta$ represents the classification accuracy improvement over the best result of the current SOTA methods in knowledge distillation.
}
\label{Table2}
\end{table}

\begin{table}[]
\centering
\scalebox{0.7}{
\begin{tabular}{cc|c|c}
\hline
 &            & top-1 acc(\%) & top-5 acc(\%) \\ \hline
\multicolumn{1}{c|}{}  &Teacher: ResNet-50    & 76.16         & 92.86         \\
\multicolumn{1}{c|}{}  & Student: MobileNet-V1 & 68.87         & 88.76         \\ \hline
\multicolumn{1}{c|}{\multirow{5}{*}{features}} & AT\cite{zagoruyko2016paying}           & 69.56         & 89.33         \\
\multicolumn{1}{c|}{}   & OFD\cite{heo2019comprehensive}          & 71.25         & 90.34         \\
\multicolumn{1}{c|}{}     & CRD\cite{tian2019contrastive}          & 71.37         & 90.41         \\
\multicolumn{1}{c|}{}   & ReviewKD\cite{chen2021distilling}     & 72.56         & 91.00         \\
\multicolumn{1}{c|}{}  & InDistill\cite{sarridis2022indistill}  & 72.52         & 91.53         \\ \hline
\multicolumn{1}{c|}{\multirow{4}{*}{logits}}   & KD\cite{hinton2015distilling}         & 70.50         & 89.80         \\
\multicolumn{1}{c|}{}   & DKD\cite{zhao2022decoupled}    & 72.50         & 91.50         \\
\multicolumn{1}{c|}{}    & CTKD\cite{li2023curriculum}    & 71.47         & 90.65         \\
\multicolumn{1}{c|}{}   & \textbf{Ours}     & \textbf{76.24}         & \textbf{93.38}         \\ 
\multicolumn{1}{c|}{}   & $\Delta$  & +3.68  & +1.85   \\
\hline
\end{tabular}}
\caption{acc refers to classification accuracy($\%$) on the ImageNet 1K dataset, top-1 and top-5 are standards for calculating classification accuracy on the validation set. $\Delta$ represents the classification accuracy improvement over the best result of the current SOTA methods in knowledge distillation.}
\label{Table3}
\end{table}

\textbf{Classification on ImageNet 1K dataset.} As shown in \cref{Table2}, we validated the capability of our method for KD where the teacher and student have similar models. Using ResNet-34 (81.39MB) as the teacher model and ResNet-18 (42.83MB) as the student model, our method demonstrates the ability to surpass the teacher model on challenging classification tasks. Moreover, compared to the best KD method, our approach achieved an additional 1.75\% improvement in classification accuracy. In \cref{Table3}, we validated the capability of our method in a cross-model knowledge distillation. Employing ResNet-50 (90.46MB) as the teacher model and MobileNet-V1 (9.01MB) as the student model, our method continues to enable the student network to surpass the teacher model in cross-model knowledge transfer. Furthermore, compared to other methods, our method achieves a substantial 3.68\% improvement in classification accuracy over the best-performing method. This improvement is particularly significant in challenging classification tasks.

\textbf{Object detection on MS-COCO dataset. }To validate the effectiveness of our method not only in classification tasks but also in object detection tasks, we applied our method on the MS-COCO dataset. The ultimate performance of object detection in this task is heavily dependent on the quality of feature extraction, particularly when dealing with the detection of numerous small objects, which significantly challenges the detector's feature extraction capability \cite{li2017mimicking,wang2019distilling}.
As shown in \cref{Table4}, our method outperforms the teacher on three metrics: $AP$, $AP_{50}$, and $AP_{70}$. This accomplishment is often challenging to achieve with conventional knowledge distillation methods.


\begin{table}[]
\centering
\scalebox{0.85}{
\begin{tabular}{c|c|c|c}
\hline
  & AP    & AP$_{50}$  & AP$_{70}$  \\ \hline
Teacher: ResNet-101   & 42.04 & 62.48 & 45.88 \\
Student: ResNet-18    & 33.26 & 53.61 & 35.26 \\ \hline
KD\cite{hinton2015distilling}    & 33.97 & 54.66 & 36.62 \\
FitNet\cite{romero2014fitnets}   & 34.13 & 54.16 & 36.71 \\
ReviewKD\cite{chen2021distilling}& 36.75 & 56.72 & 34.00 \\
InDistill\cite{sarridis2022indistill}  & 34.93 & 56.56 & 37.46 \\
DKD\cite{zhao2022decoupled}      & 35.05 & 56.60 & 37.54 \\
CTKD\cite{li2023curriculum}      & 34.56 & 55.43 & 36.91 \\
\textbf{Ours}  & \textbf{42.10}  & \textbf{62.59}   & \textbf{45.91} \\ \hline
Teacher: ResNet-50    & 40.22 & 61.02 & 43.81 \\
Student: MobileNet-V2 & 29.47 & 48.87 & 30.90 \\ \hline
KD\cite{hinton2015distilling}    & 30.13 & 50.28 & 31.35 \\
FitNet\cite{romero2014fitnets}   & 30.20 & 49.80 & 31.69 \\
ReviewKD\cite{chen2021distilling}& 33.71 & 53.15 & 36.13 \\
InDistill\cite{sarridis2022indistill}  & 32.17 & 53.49 & 34.56 \\
DKD\cite{zhao2022decoupled}      & 32.34 & 53.77 & 34.01 \\
CTKD\cite{li2023curriculum}      & 31.39 & 52.34 & 33.10 \\
\textbf{Ours}   & \textbf{41.02} & \textbf{62.13} & \textbf{44.08} \\   \hline
\end{tabular}}
\caption{Results on MS-COCO based on Faster-RCNN \cite{ren2015faster}-FPN\cite{lin2017feature}: AP evaluated on val2017.}
\label{Table4}
\end{table}

\subsection{Ablation Analysis}

\begin{table}[]
\centering
\begin{tabular}{ccccc}
\hline
\multicolumn{1}{c|}{Student}  & \multicolumn{1}{c}{EBK} & \multicolumn{1}{c|}{RBK} & \multicolumn{1}{c|}{acc(\%)} & \multicolumn{1}{c}{$\Delta$} \\ \hline
\multicolumn{5}{c}{ResNet-50 as the teacher(79.34)}      \\ \hline
\multicolumn{1}{c|}{\multirow{4}{*}{ResNet-18}}  &   &\multicolumn{1}{c|}{}   &\multicolumn{1}{c|}{71.56}   &\multicolumn{1}{c}{-}   \\
\multicolumn{1}{c|}{}   &$\surd$  &\multicolumn{1}{c|}{}    &\multicolumn{1}{c|}{79.70}    &\multicolumn{1}{c}{+8.26}     \\
\multicolumn{1}{c|}{}   &   &\multicolumn{1}{c|}{$\surd$}   &\multicolumn{1}{c|}{73.86}    &\multicolumn{1}{c}{+2.30}     \\
\multicolumn{1}{c|}{}   &$\surd$ &\multicolumn{1}{c|}{$\surd$}    &\multicolumn{1}{c|}{81.50}    & +9.94                 \\ \hline
\multicolumn{1}{c|}{\multirow{4}{*}{MobileNet-V2}} &   &\multicolumn{1}{c|}{}   &\multicolumn{1}{c|}{69.35}    & -    \\
\multicolumn{1}{c|}{}      &$\surd$  &\multicolumn{1}{c|}{}     &\multicolumn{1}{c|}{79.85}     &+10.50      \\
\multicolumn{1}{c|}{}      &   &\multicolumn{1}{c|}{$\surd$}     &\multicolumn{1}{c|}{71.27}    &+1.92       \\
\multicolumn{1}{c|}{}      &$\surd$   &\multicolumn{1}{c|}{$\surd$}   &\multicolumn{1}{c|}{80.45} &+11.10  \\ \hline           
\end{tabular}
\caption{The acc denotes the classification accuracy on the CIFAR-100 validation dataset, and the $\Delta$ represents the improvement in classification accuracy compared to the baseline KD method. The $\surd$ symbol indicates the usage of the corresponding module, while the absence of $\surd$ indicates the absence of both modules, implying the use of the basic KD method.}
\label{Table5}
\end{table}

We have validated the effectiveness of two modules for the elimination and rectification of biased knowledge, as shown in \cref{Table5}. EBK represents KD after the elimination of biased knowledge, while RBK represents KD after the rectification of biased knowledge. Validation was performed on the CIFAR-100 dataset, with the teacher model selected as ResNet-50 (79.34\%). We chose ResNet-18 (69.75\%) and MobileNet-V2 (65.40\%) with the same model structures and different as the student models, respectively. It can be observed that when using either of the two modules alone, the performance surpasses that of regular knowledge distillation. The best results are achieved when both modules are applied together in knowledge distillation. This effect is evident across both identical and different model structures.

\begin{figure}
\centering
\includegraphics[scale=0.6]{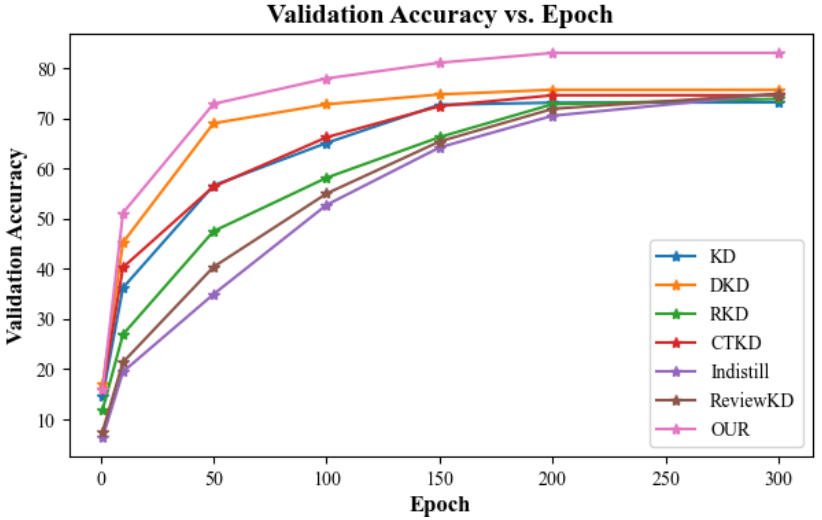}
\caption{The accuracy variation curves during the training process for several KD methods. The x-axis is training epochs, and the y-axis is the accuracy of the ResNet-34 model on the CIFAR-100 validation set.
}
\label{fig3}
\end{figure}

As shown in \cref{fig3}, our method is compared with several representative methods. It can be seen that within the same training period, our method (pink line) achieves the highest final accuracy. Our method requires fewer training epochs to reach the same accuracy level of the student model compared to others, excelling in both final accuracy and training speed. By using the approach, we achieved dynamic adjustments throughout the overall training process. This approach helps accelerate the learning process of the student network during knowledge distillation. Compared to methods that do not adopt this approach, it reduces the training time cost by 25\%.

\textbf{Wide Applicability. }Our method serves as a knowledge distillation strategy to enhance the effectiveness of existing knowledge distillation methods. As shown in \cref{Table6}, our method consistently improves both feature-based and logit-based knowledge distillation methods. The combination with the DKD\cite{zhao2022decoupled} method has achieved the best results, which is attributed to the ability of eliminating the influence of biased knowledge by our method, strengthening the transmission of right knowledge, and addressing the confusion caused by biased knowledge for students learning with the DKD method. Therefore, the combination of these two methods has yielded the most outstanding learning outcomes. In feature-based methods, the student network may be influenced by certain misleading information present in the teacher network's intermediate layer knowledge, which is challenging to effectively eliminate. Consequently, the performance improvement of such methods is not as significant as that of logit-based methods. However, feature-based methods still offer notable performance gains.

\begin{table}[]
\centering
\scalebox{0.85}{
\begin{tabular}{c|c|c|c}
\hline
                    & \multicolumn{1}{l|}{} & \multicolumn{1}{l|}{acc(\%)} & \multicolumn{1}{c}{$\Delta$} \\ \hline
                    & teacher(ResNet-50)    &  79.34        & -         \\
                    & student(ResNet-18)    &  69.75        & -         \\ \hline
\multirow{8}{*}{\rotatebox{90}{features}} 
                    & FitNet\cite{romero2014fitnets} &  70.21        & -         \\
                    & FitNet+Ours           &  72.96        & +2.75      \\ \cline{2-4} 
                    & RKD\cite{park2019relational} &  71.67        & -         \\
                    & RKD+Ours              &  74.07        & +2.40      \\ \cline{2-4} 
                    & CRD\cite{tian2019contrastive} &  72.16        & -         \\
                    & CRD+Ours              &  74.21        & +2.05      \\ \cline{2-4} 
                    & InDistill\cite{sarridis2022indistill} &  73.17        & -         \\
                    & InDistill+Ours        &  74.83        & +1.66      \\ \hline
\multirow{6}{*}{\rotatebox{90}{logits}} 
                    & KD\cite{hinton2015distilling}  &  71.56        & -         \\
                    & KD+Ours               &  78.81        & +7.25     \\ \cline{2-4} 
                    & DKD\cite{zhao2022decoupled}  &  73.97        & -         \\
                    & DKD+Ours              &  80.17        & +6.20     \\ \cline{2-4} 
                    & CTKD\cite{li2023curriculum}  &  72.29        & -         \\
                    & CTKD+Ours             &  79.00        & +6.71     \\ \hline             
\end{tabular}}
\caption{The table compares the performance of our method applied to other knowledge distillation mathod. Accuracy represents the classification accuracy on the CIFAR-100 dataset, $\Delta$ indicates the difference from the original method results, and the $+$ indicates improvement.}
\label{Table6}
\end{table}

\section{Conclusion}
We introduce a novel KD framework that tackles the challenge of biased knowledge transferring from teacher to student. This novel framework can completely eliminate the influence of biased knowledge of the teacher and substantially enhance the student by correcting knowledge through an elaborately rectifying strategy. We conduct extensive experiments on four widely used datasets and eight sets of teacher-student models, achieving SOTA results. Experimental results validate that our framework effectively boosts the performance of student, even surpassing the teacher in most scenarios. Furthermore, our framework is compatible with various existing KD methods and enhances their effectiveness.

\section*{Acknowledgements}

This work is supported by the National Natural Science
Foundation of China Excellent Young Scientists Fund
(Grant No. T2422015), the National Natural Science
Foundation of China Youth Fund (Grant No. 62202137 and 62306212) and the Marie Skłodowska-Curie Postdoctoral Individual Fellowship under Grant No. 101154277.

\bibliography{aaai25}

\end{document}


\appendix
\section{Supplementary Material}

\newtheorem{proposition}{Proposition}

\subsection{Proof}
\begin{proposition}
    The biased knowledge transmitted by the teacher has a negative influence on students' predictive outcomes.
\end{proposition}

\begin{proof}

\begin{equation}
    Loss = KL\left( T,S \right) + CE\left( Y,S \right)
\end{equation}

\begin{align}
    &min\left(Loss\right) = min\left(KL\right) + min\left(CE\right)    \\
    &KL\left(T,S\right) = \sum\limits_{i=1}^n{t_i{ln\frac{t_i}{s_i}}} = -\sum\limits_{i=1}^n{t_i{ln\frac{s_i}{t_i}}} \\
    &\because ln\left( x \right) \leq x - 1, \forall{x>0}.  \\
    &\therefore ln\left( \frac{s_i}{t_i} \right) \leq \left( \frac{s_i}{t_i} - 1\right) \\
    &\therefore -\sum\limits_{i=1}^n{t_i{ln\frac{s_i}{t_i}}} \geq  -\sum\limits_{i=1}^n{t_i\left({\frac{s_i}{t_i}} - 1\right)}  \\
    &\because -\sum\limits_{i=1}^n{t_i\left({\frac{s_i}{t_i}} - 1\right)} = -\sum\limits_{i=1}^n{t_i-{s_i}} \\
    &\because \sum\limits_{i=1}^n{t_i}=1, \sum\limits_{i=1}^n{s_i}=1.  \\
    &\therefore -\sum\limits_{i=1}^n{t_i\left({\frac{s_i}{t_i}} - 1\right)} = -\sum\limits_{i=1}^n{t_i-{s_i}}=0  \\
    &\therefore -\sum\limits_{i=1}^n{t_i{ln\frac{s_i}{t_i}}} \geq 0  
\end{align}

    when $min\left( KL \right) \to 0, \frac{s_i}{t_i} \to 1$, Thus $s_i \cong t_i$

\begin{equation}
    CE\left( Y,S \right) = -\sum_{i=1}^n {y_i ln s_i},~~ s_{i} = \frac{e^{z_{i} } }{\sum_{i=1}^{n} e^{z_{i} }}
\end{equation}

\begin{equation}
    \frac{\partial \left ( CE \right ) }{\partial z_i } = \frac{\partial \left ( CE \right ) }{\partial s_i } \times \frac{\partial s_i}{\partial z_i } 
\end{equation}    

\begin{equation}
    \frac{\partial \left ( CE \right ) }{\partial s_i } = - {\sum_{i=0}^{n}} \frac{y_i}{s_i} 
\end{equation}

\begin{equation}
\because
    \begin{cases}
        y_{i=c} = 1\\
        y_{i\ne c} = 0
    \end{cases}
\end{equation}
where $y$ is a set of binary vectors forming labels, $y_c$ is encoded as 1 to represent the corresponding category, and $y_{i\ne c}$ is encoded as 0.

\begin{equation}
    \therefore
    \frac{\partial \left ( CE \right ) }{\partial s_i} =-\frac{1}{s_c} , \frac{\partial \left ( CE \right ) }{\partial z_i} = -\frac{1}{s_c}\times \frac{\partial s_c}{\partial z_i} 
\end{equation}

\begin{equation}
    s_c = \frac{e^{z_c}}{\sum_{i=1}^{n} e^{z_i}} ,
    \frac{\partial s_c}{\partial z_i} = \frac{\frac{\partial e^{z_c} }{\partial e^{z_i}} \sum_{i=1}^{n} e^{z_i} -e^{z_c}e^{z_i}}{{\left ( \sum_{i=1}^{n} e^{z_i} \right ) }^2 } 
\end{equation}
when $c = i$,

\begin{align}
    \frac{\partial e^{z_c} }{\partial e^{z_i}} &= e^{z_i} ,
    \frac{\partial s_c}{\partial z_i} = \frac{e^{z_i} \sum_{i=1}^{n} e^{z_i} -e^{z_c}e^{z_i}}{{\left ( \sum_{i=1}^{n} e^{z_i} \right ) }^2 }   \\
    \frac{\partial \left ( CE \right ) }{\partial z_i} &= -\frac{\sum_{i=1}^{n} e^{z_i}}{e^{z_i}} \times \frac{e^{z_i} \sum_{i=1}^{n} e^{z_i} -e^{z_c}e^{z_i}}{{\left ( \sum_{i=1}^{n} e^{z_i} \right ) }^2 }  \\
    &= \frac{e^{z_i}}{\sum_{i=1}^{n} e^{z_i}} -1 = s_i - 1
\end{align}

when $c \ne i$,

\begin{align}
    \frac{\partial e^{z_c} }{\partial e^{z_i}} &= 0 ,
    \frac{\partial s_c}{\partial z_i} = \frac{-e^{z_c}\times e^{z_i}}{{\left ( \sum_{i=1}^{n} e^{z_i} \right ) }^2 }   \\
    \frac{\partial \left ( CE \right ) }{\partial z_i} &= -\frac{\sum_{i=1}^{n} e^{z_i}}{e^{z_i}} \times \frac{-e^{z_c}\times e^{z_i}}{{\left ( \sum_{i=1}^{n} e^{z_i} \right ) }^2 }  \\
    &= \frac{e^{z_i}}{\sum_{i=1}^{n} e^{z_i}} = s_i 
\end{align}

thus
\begin{equation}
    \frac{\partial \left ( CE \right ) }{\partial z_i} = \begin{cases}
         s_i -1, c=i\\
         s_i~~~~~~~, c \ne i
    \end{cases}
\end{equation}

let
\begin{equation}
    \frac{\partial \left ( CE \right ) }{\partial z_i} =0 \Rightarrow 
    \begin{cases}
         s_{i=c} = 1\\
         s_{i \ne c} = 0
    \end{cases} 
\end{equation}

Thus, minimizing $\left( CE \right)$ drives the correct class $s_c \to$ 1, while the other classes $s_{i \ne c} \to$ 0.

Let $a$ and $b$ represent two classes, where $t_a$ and $t_b$ are the teacher's predicted probabilities, $s_a$ and $s_b$ are the student's predicted probabilities, and $y_a$ and $y_b$ are the corresponding class labels. Where, assuming $y_a = 1$ and $y_b = 0$, when $min\left( loss \right)$, $s_a \to 1$ and $s_b \to 0$ simultaneously, $s_a \to t_a$ and $s_b \to t_b$. 
\begin{equation}
    min\left(Loss\right) \Rightarrow
    \begin{cases}
     s_a \to 1, s_a \to t_a\\
     s_b \to 0, s_b \to t_b
    \end{cases}
\end{equation}

If the teacher's predictions are consistent with the labels, i.e., $t_a\to 1$, $t_b\to 0$, when we minimize the $Loss$ for regression, $min(CE)$ will yield an $s_a$ value close to the label($s_a \to y_a$). At this point, through learning from the teacher($t_a$), $min(KL)$ will get $s_a$ closer to $t_a$($s_a\to t_a$). Ultimately, driven by minimizing the $Loss$, the value of $s_a$ will lie between $t_a$ and $y_a$($t_a<s_a<1$). Similarly, $0<s_b<t_b$ also holds. Through learning from the teacher's knowledge, this ensures that the student better distinguishes between correct and incorrect classes. Such teacher knowledge is beneficial for the student.

However, if the teacher's predictions are inconsistent with the labels, i.e, $t_a\to 0$, $t_b\to 1$, during the process of obtaining the most appropriate values for $s_a$ and $s_b$ through regression, $min(CE)$ will make $s_a$ tend towards $y_a$, while $min(KL)$ will make $s_a$ tend towards $t_a$. At this point, since $t_a$ tends towards 0, it causes $s_a$, which should tend towards $y_a$, to be influenced by $t_a$ and deviate away from $y_a$. Similarly, due to $t_b$ tending towards 1, $s_b$, which should tend towards $y_b$, is influenced by $t_b$ and deviate away from $y_b$. Ultimately, this hinders the student's ability to differentiate between correct and incorrect classes. Such teacher biased knowledge is harmful for the student.


\end{proof}

\subsection{Extended Experiments}

As shown in \cref{tab1}, we conducted comparative experiments on the CIFAR-100 dataset to observe the changes in classification accuracy of the student network during the knowledge distillation process with respect to the number of training epochs. In these experiments, "normal" refers to the standard training method, while "ours" refers to the dynamically adjusted training method. It can be seen that our method quickly improves accuracy within the first 100 epochs, achieving better results than the normal method after 200 epochs. After 200 epochs, the performance of our method has already surpassed that of the standard method by the time 300 epochs are reached.

During the overall KD training process, the knowledge transferred by the teacher can be classified into right and bias knowledge. For the student, experience in easy tasks comes from the teacher's right knowledge, while experience in hard tasks comes from the knowledge obtained after rectifying bias. Therefore, during the early stages of student learning, the student is more likely to successfully learn these easy tasks. Once the student has acquired a certain level of capability, focusing on bias knowledge after rectification will further enhance the student's overall ability. Thus, our dynamic adjustment method allows the student to achieve optimal performance while reducing the training period.

\begin{table}[b]
\centering
\scalebox{0.8}{
\begin{tabular}{c|ccccccc}
\multirow{2}{*}{} & \multicolumn{7}{c}{T:ResNet-101 (81.87)}              \\
                  & \multicolumn{7}{c}{S:ResNet-34 (72.50)}               \\ \hline
epoch             & 1     & 10    & 50    & 100   & 150   & 200   & 300   \\ \hline
normal            & 10.99 & 26.65 & 46.56 & 59.41 & 69.01 & 74.87 & 84.02 \\ \hline
ours              & 16.11 & 51.15 & 72.85 & 77.92 & 81.05 & 84.17 & -     \\ \hline
\multirow{2}{*}{} & \multicolumn{7}{c}{T:ResNet-50 (79.34)}               \\
                  & \multicolumn{7}{c}{S:MobileNet-V2 (65.40)}            \\ \hline
epoch             & 1     & 10    & 50    & 100   & 150   & 200   & 300   \\ \hline
normal            & 12.27 & 34.99 & 54.01 & 61.84 & 66.35 & 70.56 & 76.83 \\ \hline
ours              & 14.36 & 40.30 & 59.79 & 69.08 & 75.22 & 80.09 & -    
\end{tabular}}
\caption{Comparison of accuracy over different training epochs between the dynamically adjusted training method and the normal training method for classification tasks.}
\label{tab1}
\end{table}

We have added experimental validation on the MiT\cite{NEURIPS2021_64f1f27b} model to demonstrate the effectiveness of our method on Transformer architectures (MiT is a variant of ViT\cite{dosovitskiy2020image} used for segmentation tasks). 
As demonstrated in \cref{tab2}, our approach significantly enhances the performance of student models when conducting knowledge distillation (KD) on models based on the Transformer architecture, enabling them to generally surpass the performance levels of their teacher models. This result indicates that our bias-mitigation strategy is not only applicable to a variety of existing models but also effectively improves the ultimate performance of current knowledge distillation methods.


\begin{table}[]
\centering
\begin{tabular}{c|c|c}
         & mIoU (\%)  & mIoU+ (\%) \\ \hline
T:MiT-B4 & 80.23 & -     \\
S:MiT-B1 & 75.30 & -     \\ \hline
SKD\cite{liu2020structured}  & 76.18 & 80.23 \\ \hline
IFVD\cite{wang2020intra}     & 76.37 & 80.29 \\ \hline
CWD\cite{Shu_2021_ICCV}   & 76.39 & 80.37 \\ \hline
DSD\cite{9444191}      & 76.89 & 80.94 \\ \hline
CIRKD\cite{yang2022cross}  & 77.01 & 81.01
\end{tabular}
\caption{
Performance of various knowledge distillation methods on the Cityscapes dataset for semantic segmentation tasks. MiT-B4 as teacher and MiT-B1 as student. The 'mIoU+' symbol indicates that the method has been improved using our proposed approach.}
\label{tab2}
\end{table}

We have also conducted experiments on the YOLOv8 model to evaluate the effectiveness of our KD method on single-stage, anchor-free object detection models. As shown in \cref{tab3}, our method outperforms other approaches across all metrics, demonstrating its broad applicability in object detection tasks, especially for single-stage anchor-free models.


\begin{table}[]
\centering
\scalebox{0.9}{
\begin{tabular}{c|c|c}
           & mAP$_{50}$ & mAP$_{50-95}$ \\ \hline
T: yolov8-m & 61.7  & 49.9     \\
S: yolov8-n & 52.1  & 37.1     \\ \hline
KD\cite{hinton2015distilling} & 55.9  & 37.8  \\ \hline
FitNet\cite{romero2014fitnets} & 54.1  & 38.0  \\ \hline
DKD\cite{zhao2022decoupled}  & 58.9  & 45.7  \\ \hline
CTKD\cite{li2023curriculum}  & 57.7  & 43.6  \\ \hline
Ours       & 62.0  & 50.1    
\end{tabular}}
\caption{Performance comparison of different knowledge distillation methods on YOLOv8 for single-stage anchor-free object detection. yolov8-m as teacher; yolov8-n as student; mAP$_{50}$: mean Average Precision at IoU=50; mAP$_{50-95}$: mean Average Precision averaged across IoU thresholds from 0.5 to 0.95.
}
\label{tab3}
\end{table}

In the Fig.3 of the main text, Step b although the two incorrect categories ($t_a$ and $t_b$) can be rectified, directly using b for knowledge distillation (KD) would weaken the student model's ability to learn the $t_o$ category, since the probabilities are not equal to 1 (see equations 5 and 6). Therefore, we propose Step c, which successfully corrects ta and tb without affecting the $t_o$ category, ensuring that the student model can more accurately learn the distribution of all categories. We present the ablation study results of methods b and c in the \cref{tab4} to visually demonstrate the effectiveness of Step c.

\begin{table}[]
\centering
\scalebox{0.9}{
\begin{tabular}{c|c|c|c}
  & T:RN-50(79.34) & T:RN-101(81.87) & T:RN-50(79.34) \\
  & S:RN-18(69.75) & S:RN-34(72.50)  & S:MN-V2(65.40) \\ \hline
b & 71.65(+1.90)   & 74.03(+1.53)    & 66.48(+1.08)   \\ \hline
c & 71.93(+2.18)   & 74.58(+2.08)    & 67.61(+2.21)  
\end{tabular}}
\caption{Comparison of the classification performance between Step b and Step c on the CIFAR-100 dataset. T and S represent the teacher and student models, respectively.}
\label{tab4}
\end{table}

\newpage

\bibliography{aaai25}